\pdfoutput=1
\documentclass[11pt]{article}

\usepackage[final]{acl}

\usepackage{times}
\usepackage{latexsym}
\usepackage{tempora} % this supports Cyrillic
\usepackage{newtxmath}
\usepackage[T1]{fontenc}
\usepackage{tcolorbox}
\usepackage[utf8]{inputenc}
\usepackage{mdframed}
\usepackage{microtype}
\usepackage{tabularx}
\usepackage{inconsolata}

\usepackage{graphicx}
\usepackage{booktabs}
\usepackage{xcolor}
\usepackage{colortbl}
\usepackage{siunitx}
\usepackage{multirow}
\usepackage{epigraph}
\usepackage{subcaption}
\usepackage[polish, ukrainian, russian, german, dutch, english]{babel}

\definecolor{lightblue}{RGB}{235,245,255}
\definecolor{lightyellow}{RGB}{255,250,235}
\definecolor{bestcolor}{RGB}{200,230,200}

\title{It's Not a Walk in the Park! \\ 
Challenges of Idiom Translation in Speech-to-text Systems}

\author{
Iuliia Zaitova\textsuperscript{1}, Badr M. Abdullah\textsuperscript{1}, Wei Xue\textsuperscript{2,1} \\
{\bf Dietrich Klakow\textsuperscript{1}}, {\bf Bernd Möbius\textsuperscript{1}}, {\bf Tania Avgustinova\textsuperscript{1}} \\
\textsuperscript{1}Saarland University, Germany \\
\textsuperscript{2}Xi'an Jiaotong University, China \\
\texttt{izaitova@lsv.uni-saarland.de}
}

\begin{document}
\maketitle
\begin{abstract}
Idioms are defined as a group of words with a figurative meaning not deducible from their individual components. Although modern machine translation systems have made remarkable progress, translating idioms remains a major challenge, especially for speech-to-text systems, where research on this topic is notably sparse. In this paper, we systematically evaluate idiom translation as compared to conventional news translation in both text-to-text machine translation (MT) and speech-to-text translation (SLT) systems across two language pairs (German to English, Russian to English). We compare state-of-the-art end-to-end SLT systems (SeamlessM4T SLT-to-text, Whisper Large v3) with MT systems (SeamlessM4T SLT-to-text, No Language Left Behind), Large Language Models (DeepSeek, LLaMA) and cascaded alternatives. Our results reveal that SLT systems experience a pronounced performance drop on idiomatic data, often reverting to literal translations even in higher layers, whereas MT systems and Large Language Models demonstrate better handling of idioms. These findings underscore the need for idiom-specific strategies and improved internal representations in SLT architectures.

\end{abstract}

\section{Introduction}
\epigraph{
  ``The difference between the right word and the almost right word is really a large matter – it’s the difference between lightning and a lightning bug.''
}{
  ---Mark Twain
}

Imagine explaining to someone unfamiliar with English that it is \emph{“raining cats and dogs”} or that you are feeling \emph{“under the weather.”} Although idioms carry meanings that cannot be derived from the meaning of individual words alone, humans can easily interpret them by relying on context and cultural knowledge. However, machine translation systems often produce literal, incorrect or nonsensical translations \cite{dankers-etal-2022-transformer, baziotis-etal-2023-automatic, rambelli-etal-2023-frequent, tian-etal-2023-idioms}.

Prior work has extensively examined idiom translation in text-based machine translation (MT) systems \cite{boisson-etal-2022-cardiffnlp, avram-etal-2023-romanian, liu-etal-2023-crossing, bui-savary-2024-cross}, yet the topic of idioms in speech translation has received comparatively little attention. Despite the success of speech translation systems such as SeamlessM4T \cite{communication2023seamlessmultilingualexpressivestreaming,Barrault2025} and Whisper \cite{radford2022robustspeechrecognitionlargescale}, which achieve state-of-the-art results across many languages and acoustic conditions, speech translation systems might be particularly prone to failing on idiomatic content due to the additional complexity of integrating acoustic, syntactic, and semantic information. Understanding if and why such failures occur is essential to further improving speech-to-text translation (SLT) systems.

\begin{figure}[t]
    \centering
        \includegraphics[width=\linewidth]{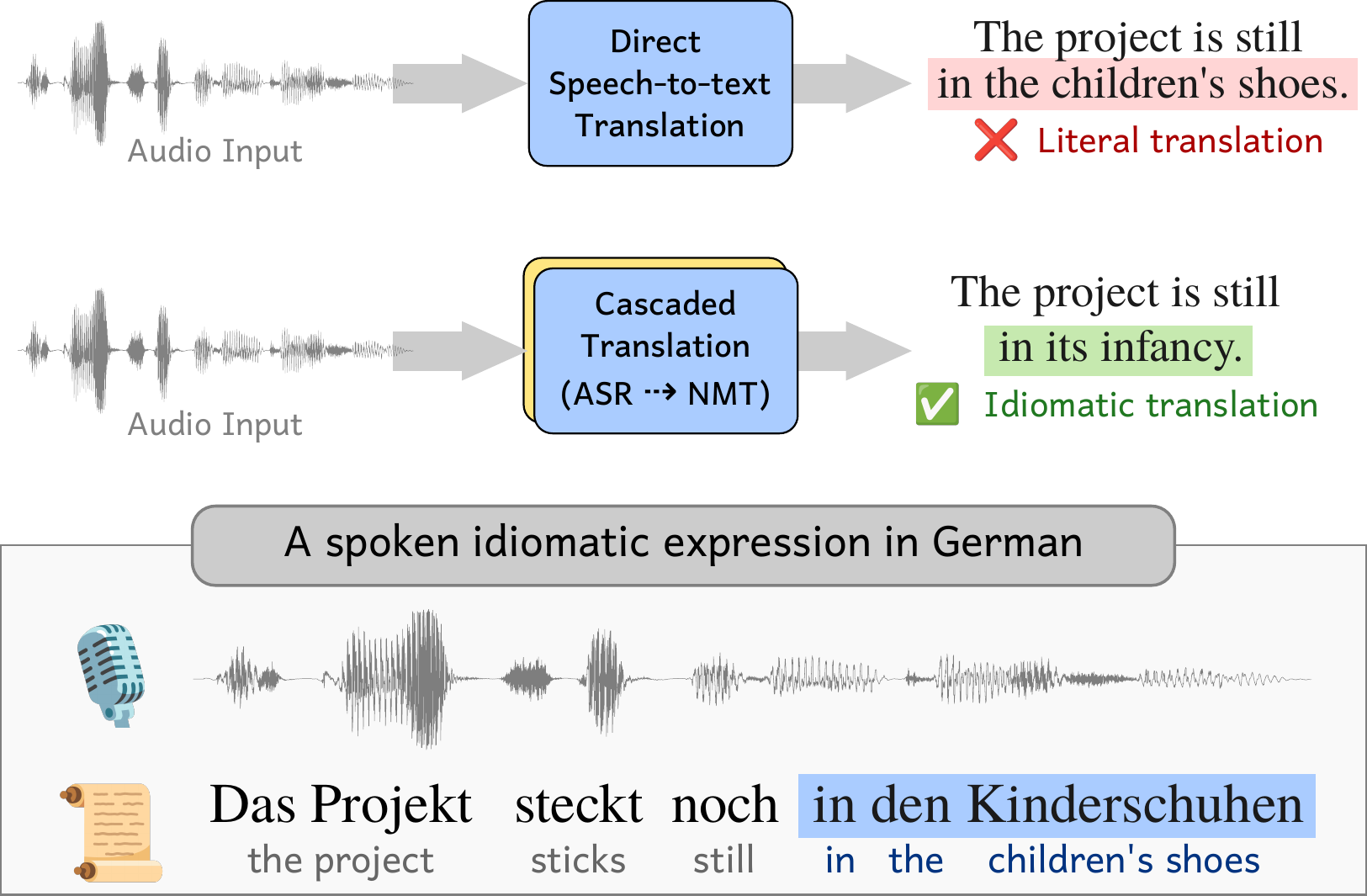}
        \caption{An illustrated example of translating a spoken idiomatic expression. The German idiom \emph{“in den Kinderschuhen”}---literally translates to \emph{“in children's shoes”}---means something is in its beginning stages, equivalent to the English \emph{“in its infancy.”} In this paper, we systematically assess the performance of two modes of spoken language translation for idiom translation: (1) direct speech-to-text translation, and (2) cascaded speech translation whereby the audio is first transcribed by a ASR system followed by a text-based machine translation. }
        \label{fig:fig_0}
\end{figure}

In this paper, we provide the first systematic comparison of idiom translation in MT, general purpose Large Language Models (LLMs), and SLT for German$\rightarrow$English and Russian$\rightarrow$English language pairs. We investigate:
\begin{itemize}
    \item The relative performance of end-to-end SLT (SeamlessM4T for audio, Whisper Large v3) vs.\ MT (SeamlessM4T for text, No Language Left Behind), general-purpose LLMs  (LLaMA 3, DeepSeek-v3), and cascaded approaches.
    \item How these systems handle idiomatic and news data, as measured by both COMET \cite{rei-etal-2020-comet} and human annotation.
    \item Layer-wise performance of MT and SLT systems via DecoderLens analysis \cite{langedijk2024decoderlenslayerwiseinterpretationencoderdecoder} to pinpoint how and at which encoder layers these systems fail on idioms.
\end{itemize}

Our experiments reveal that SLT significantly underperforms MT and Large Language Models (LLMs) on idiomatic data, even though they perform competitively on conventional news text. We make our code, evaluation datasets, and their annotated subsets publicly available at \url{https://github.com/IuliiaZaitova/idiom_s2t}.

\section{Related Work}
\label{sec:related_work}

\paragraph{Idiom Translation in text-based systems.}
The difficulty of translating and handling idioms has been extensively studied in MT systems and LLMs.
For instance, \citet{dankers-etal-2022-transformer} and \citet{baziotis-etal-2023-automatic} explored how Transformer architectures handle figurative language, identifying a tendency to produce literal translation.

Strategies such as fine-tuning on idiom-focused parallel data
\cite{boisson-etal-2022-cardiffnlp, avram-etal-2023-romanian} have shown promising improvements in idiom translation accuracy, though translation systems remain vulnerable to varied contexts and domains.

\paragraph{SLT Systems.}
SLT has seen significant advances with recent end-to-end architectures such as Whisper~\cite{radford2022robustspeechrecognitionlargescale} and SeamlessM4T~\cite{communication2023seamlessmultilingualexpressivestreaming,Barrault2025}.
Earlier SLT research often relied on cascaded approaches,
combining an automatic speech recognition (ASR) module with a separate MT system \cite{niehues2018lowlatencyneuralspeechtranslation, IRANZOSANCHEZ2021303}. Recently, cascaded speech-to-text translation models
have encountered criticism due to an intrinsic
shortcoming of 'error propagation'. Techniques were proposed to mitigate this shortcoming and enhance the accuracy of the translation in cascaded systems~\cite{min2025endtoendoverkillrethinkingcascaded}.
However, the IWSLT 2023 Evaluation Campaign \cite{agrawal-etal-2023-findings} still notes that cascaded approaches remain competitive in certain scenarios. These systems often outperform end-to-end systems when leveraging high-resource ASR and MT components, especially for languages with limited training data for direct SLT. 

\paragraph{Evaluation of Figurative Language Translation.}
\citealp{song-xu-2024-benchmarking} explore which automatic metrics work best for evaluating multiword expressions (MWEs) and figurative language in translation.
They conclude that surface-level string metrics like BLEU~\cite{papineni-etal-2002-bleu} often fail to capture nuanced meaning shifts in idiomatic data, whereas
semantic metrics like COMET~\cite{rei-etal-2020-comet} correlate more reliably with human judgments of MWE translation quality.

\paragraph{Interpretability and Layer-wise Analysis.}
In parallel with improvements in model performance, interpretability methods seek to reveal \emph{how} and \emph{where} complex systems process inputs.
\citet{voita-etal-2019-analyzing} and \citet{clark-etal-2019-bert} examine attention heads in Transformer models, showing that syntactic and semantic information is often distributed across multiple layers.
More recently, \citet{langedijk2024decoderlenslayerwiseinterpretationencoderdecoder} proposed DecoderLens analysis, which replaces a model’s final encoder output with intermediate layer representations, translating them to human-readable text.
This method offers deeper insight into how the output evolves throughout the encoding process, which is particularly useful for diagnosing issues of incorrect translation.

\section{Methodology}

\subsection{Task and Scope}
Idioms present unique challenges in translation due to their non-literal nature, which often requires contextual and cultural understanding. We focus on translating idiomatic and, for contrast, conventional news datasets in two language pairs (German$\rightarrow$English, Russian$\rightarrow$English) across speech and text modalities. 

\subsection{Systems Evaluated}

\paragraph{MT Systems}
\begin{enumerate}
    \item  SeamlessM4T (text-to-text) with version \texttt{facebook/seamless-m4t-v2-large}: A state-of-the-art multilingual MT system capable of direct text-to-text translation across multiple languages.
\item No Language Left Behind (NLLB) with version \texttt{facebook/nllb-200-3.3B}: A system developed for enhancing translation quality in low-resource languages, capable of translating over 202 different languages with state-of-the-art results~\cite{nllbteam2022languageleftbehindscaling}.
\end{enumerate}

\paragraph{Large Language Models (LLMs)}
\begin{enumerate}
    \item LLaMA~3 models fine-tuned for specific languages:
    (a)   {\small\texttt\allowbreak{{IlyaGusev/saiga\_LLaMA3\_8b}}}~\cite{saiga_LLaMA3_8b} fine-tuned for Russian, and (b)  {\small\texttt\allowbreak{VAGOsolutions/LLaMA-3-SauerkrautLM-8b-Instruct}}~\cite{LLaMA3_sauerkrautlm_8b} fine-tuned for German.

    \item DeepSeek-V3~\cite{deepseekai2025deepseekv3technicalreport}: A multilingual LLM optimized for translation, reasoning, and code generation tasks. It tops the leaderboard among open-source models.
\end{enumerate}

\paragraph{LLM Prompts} To ensure transparency, we include the prompts used to produce translation to English by LLaMA and DeepSeek models in Appendix~\ref{sec:appendix_A}.

\paragraph{SLT Systems} 
\begin{enumerate}
\item SeamlessM4T (speech-to-text) with version \texttt{facebook/seamless-m4t-v2-large}: An end-to-end multilingual system capable of translating speech inputs into text.
\item Whisper Large v3 with version \texttt{openai/whisper-large-v3} (Whisper): A highly robust speech recognition and translation model with 1.55 billion parameters, designed to handle diverse languages and acoustic conditions.
\end{enumerate}
\paragraph{Cascaded Systems}
We formed cascaded systems by feeding audio inputs (16kHz mono WAV) into either SeamlessM4T or Whisper for ASR, then passing their transcriptions into each MT system and LLM. The transcribed text’s capitalization and punctuation was retained.

\begin{table*}[ht]
\centering
\small
\renewcommand{\arraystretch}{1.25}
\setlength{\tabcolsep}{7pt}
\definecolor{LightGray}{RGB}{240,240,240}
\begin{tabular}{@{}>{\raggedright}p{3.2cm}p{4.8cm}p{5.5cm}@{}}
\toprule
\rowcolor{LightGray}
\textbf{Category} & \textbf{Description} & \textbf{Example (De)} \\
\midrule

\rowcolor{LightGray}
\multirow{2}{*}{\textbf{Correct}} 
& \textit{Idiomatic} $\dagger$: Preserves figurative meaning & 
\textit{Es ist mir wurst} $\rightarrow$ \textit{I couldn't care less} \\
\cmidrule(lr){2-3}
& \textit{Paraphrase} $\dagger$: Literal conversion with meaning & 
\textit{Es ist mir wurst} $\rightarrow$ \textit{It doesn't matter} \\ 
\midrule

\rowcolor{LightGray}\textbf{Partially Correct} & Core meaning with minor errors; more than 50\% of the sentence is translated correctly & 
\textit{Es ist mir wurst} $\rightarrow$ \textit{It matters to me} \\ 
\midrule

\textbf{Literal Translation} $\dagger$ & Word-for-word idiom translation that loses the idiomatic meaning; the sentence translation otherwise correct & 
\textit{Es ist mir wurst} $\rightarrow$ \textit{It is sausage} \\ 
\midrule

\rowcolor{LightGray}\textbf{Incorrect (Relevant)} & Addresses the same topic but misrepresents critical information; less than 50\% of the sentence is translated correctly & 
\textit{Es ist mir wurst} $\rightarrow$ \textit{I want to go} \\ 
\midrule

\textbf{Incorrect (Hallucination)} & Fabricated unrelated content & 
\textit{Es ist mir wurst} $\rightarrow$ \textit{ I'm not a child} \\ 
\midrule

\rowcolor{LightGray}\textbf{Empty/Ellipsis} & Missing/empty output & 
\textit{Es ist mir wurst} $\rightarrow$ \textit{,,,,,} \\ 
\bottomrule
\end{tabular}
\vspace{0.5em}
\caption{Annotation scheme for manual translation evaluation. $\dagger$ marks categories specific to idiom evaluation. The German phrase \emph{'Es ist mir wurst'} is correctly translated to English as \emph{'I couldn't care less'.}}
\label{tab:annotation_scheme}
\end{table*}

\subsection{Evaluation Datasets}
\subsubsection{Conventional News Corpus}
To evaluate general translation performance, we used the professionally translated \emph{News Commentary} parallel corpus\footnote{\url{https://metatext.io/datasets/news-commentary-parallel-corpus}}. This dataset includes formal, well-structured news text in political and economic domain with minimal use of figurative language, making it ideal as a baseline for general translation performance. By providing consistent and straightforward content, the \emph{News Commentary} corpus allows us to contrast the performance of translation systems under conventional conditions with their ability to handle idiomatic data. To perform our evaluation, we randomly selected 250 sentences from the News Commentary corpus for both language pairs. Examples from the \emph{News Commentary} corpus are shown below:
\begin{quote}
\textbf{Russian:} \selectlanguage{russian}{\emph{Что же может оправдать очередной значительный рост цен на золото, начиная с сегодняшнего дня?}} \selectlanguage{english}{(Eng. trans.: So what could justify another huge increase in gold prices from here?)}
\end{quote}\begin{quote} \textbf{German:} \emph{Damals lag Gold bei 850 Dollar, also in heutigem Geldwert um einiges über 2.000 Dollar.} (Eng. trans.:  Back then, gold hit \$850, or well over \$2,000 in today’s dollars.)
\end{quote}

\subsubsection{Idiomatic Corpus}
Idiomatic data used for evaluation is sourced from the \emph{Idioms-InContext-MT} dataset~\cite{stap2024finetuningparadoxboostingtranslation}\footnote{\url{https://github.com/amazon-science/idioms-incontext-mt}}. From the 1,000 examples available in the dataset per language pair, we manually selected 250 idioms that require non-literal translation to preserve their figurative meaning. For instance:
\begin{quote} 
\textbf{German:} \emph{\textbf{Es ist mir wurst}, wenn du nicht kommst.} (literally: \emph{It is sausage to me if you don't come.}, meaning: \emph{Eng. trans.: I couldn't care less if you don't come.})
\end{quote}

\begin{quote}\textbf{Russian:} — \selectlanguage{russian}{Ну да! Мы с тобой — \textbf{два сапога пара}! — охотно согласился Шурик.} \selectlanguage{english}{(literally: \emph{Well, yes! You and me are like two boots of a pair!, Shurik happily agreed.}, meaning: \emph{Eng. trans.: Well, yes! You and me are like two peas in a pod!, Shurik happily agreed.}})\end{quote}

We excluded idioms whose figurative meaning is preserved in a literal translation. For example:
\begin{quote}\textbf{Russian:} \selectlanguage{russian}{\emph{Они и \textbf{мухи не обидят.}}} \selectlanguage{english} (literally: \emph{They wouldn't hurt a fly.}) \end{quote} 
\begin{quote}\textbf{German:} \emph{Als er die Nachricht hörte, \textbf{brach es ihm das Herz.}} (literally: \emph{When he heard the news, it broke his heart.})\end{quote} 
This selection process ensures the focus remains on idioms that challenge machine translation systems, allowing us to evaluate their ability to translate idiom figuratively.

%\subsection{Speech Data}
%\begin{figure}[t]
%    \centering
%    \begin{subfigure}{0.48\textwidth}
%        \includegraphics[width=\linewidth]{images/de_performance_comparison.pdf}
%        \caption{COMET scores for German}
%        \label{fig:comet_german}
%    \end{subfigure}
%    \hfill
%    \begin{subfigure}{0.48\textwidth}
%        \includegraphics[width=\linewidth]{images/ru_performance_comparison.pdf}
%        \caption{COMET scores for Russian}
%        \label{fig:comet_russian}
%    \end{subfigure}
%    \caption{Comparison of COMET scores for German and Russian translations. MT -- machine translation, SLT -- spoken language translation, ASR -- automatic speech recognition.}
%    \label{fig:comet_comparison}
%\end{figure}

\begin{figure*}[t]
  \includegraphics[width=\textwidth]{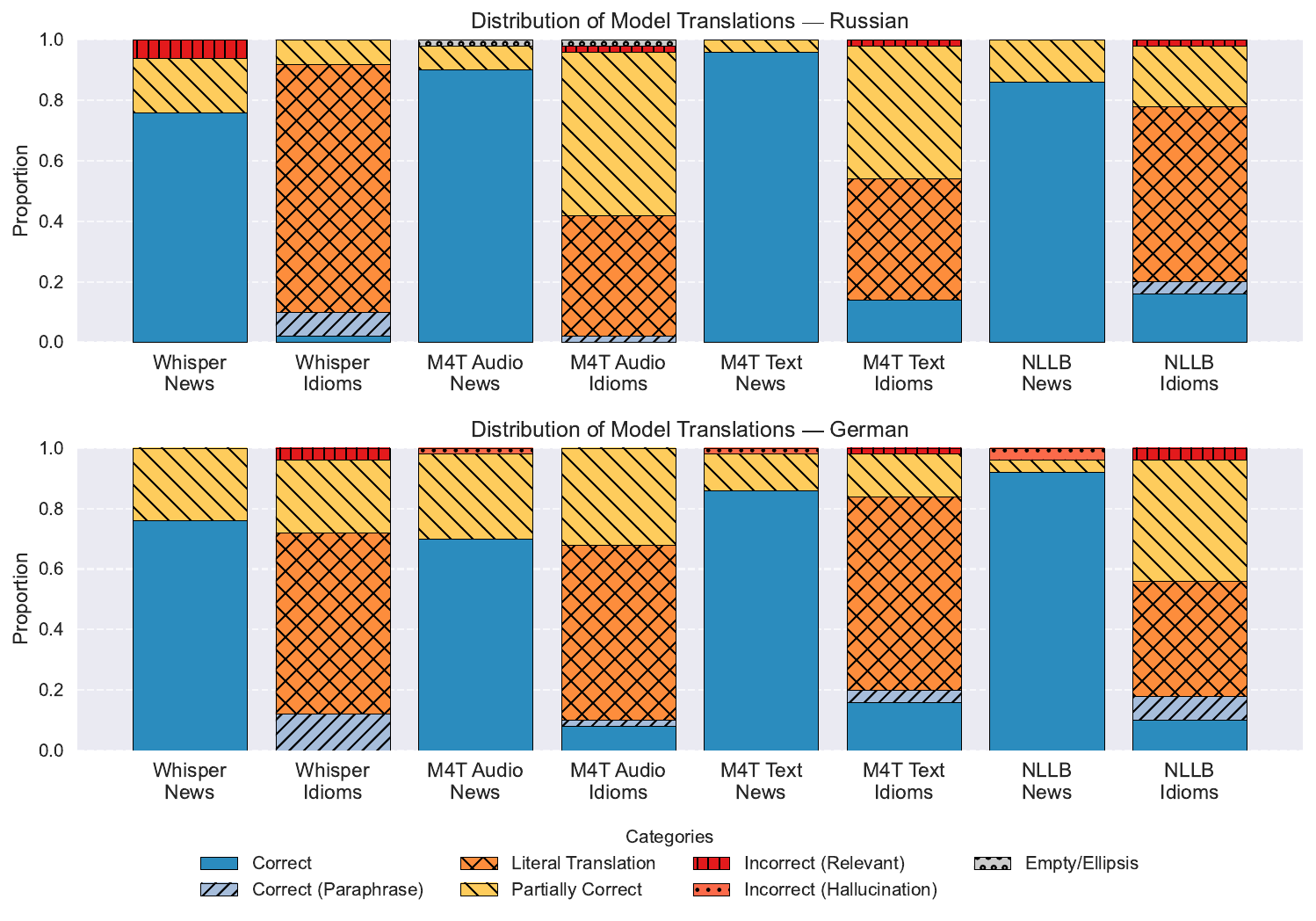}
  \caption{Distribution of translation output categories across models for German$\rightarrow$English and Russian$\rightarrow$English translation. Each bar represents a model's output distribution on either news or idiomatic test sets. Speech-to-text translation systems mostly show lower proportions of correct translations for idioms compared to text-to-text translation systems, indicating a particular challenge of idiom translation in speech-to-text systems.}
  \label{fig:output_categories}
\end{figure*}

\begin{figure*}[t]
  \includegraphics[width=\textwidth]
  {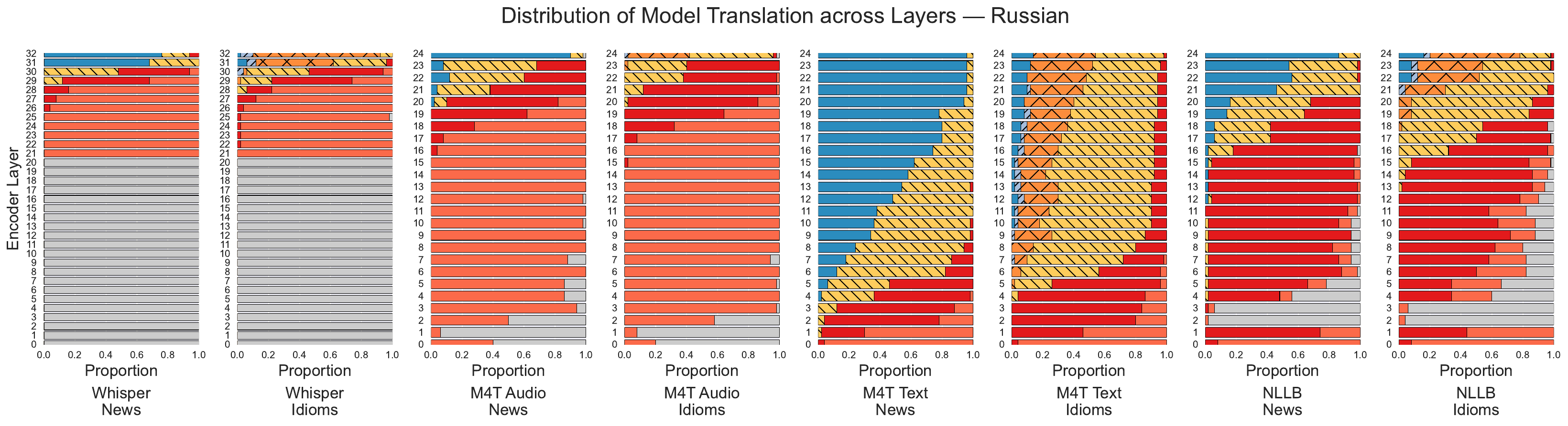}
  \includegraphics[width=\textwidth]{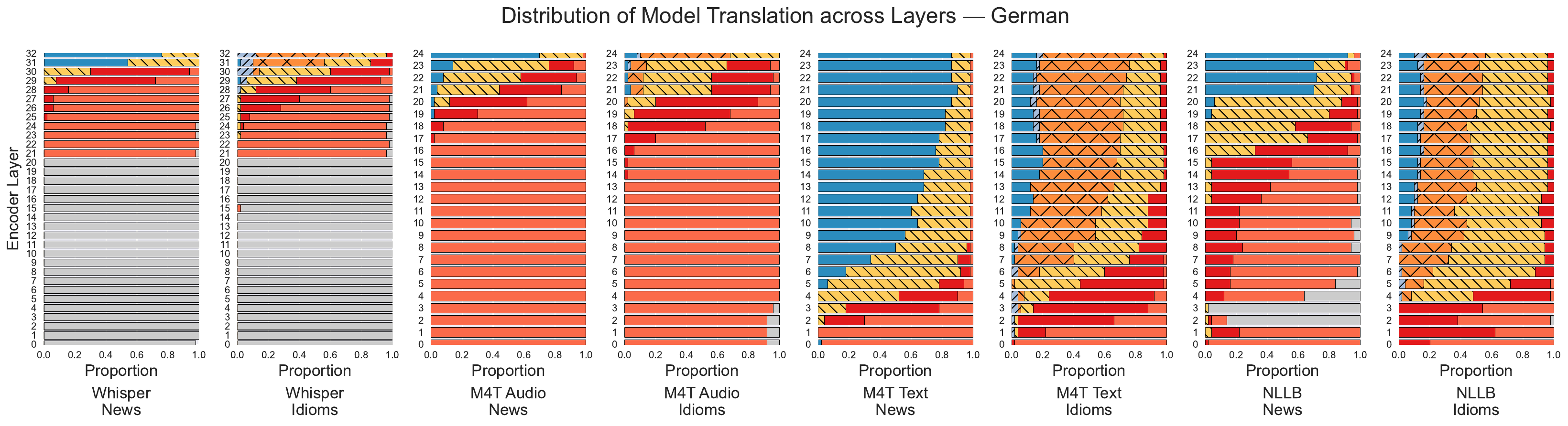}
  \includegraphics[width=\textwidth]{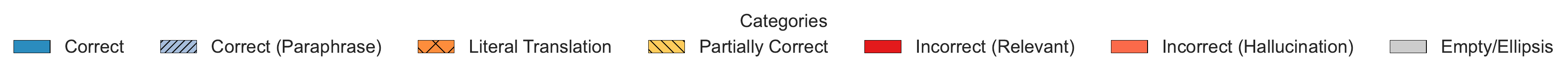}
  \caption{Distribution of translation error categories across encoder layers for German$\rightarrow$English and Russian$\rightarrow$English translation. Each subplot shows the evolution of translation quality through different encoder layers for a specific model and domain (news vs. idioms). The x-axis shows the proportion of translations falling into each category, and y-axis represents encoder layers.}
  \label{fig:decoderlens_categories}
\end{figure*}

To enable SLT evaluation, we synthesized audio for all text segments using Microsoft Edge voice services\footnote{\url{https://www.microsoft.com/edge}}, which employs neural text-to-speech (TTS) architectures comparable to state-of-the-art systems. Synthesizing speech for text-based datasets is a widely used practice in translation research~\cite{jia2019leveragingweaklysuperviseddata, moslem2024leveragingsyntheticaudiodata, bamfo-odoom-etal-2024-synthetic}. 

 While synthetic speech may have minor deviations in prosody or emphasis~\cite{wester2016evaluating, Chan2024ExploringTA}, such factors are secondary in idiomaticity-centered MT evaluation. Modern TTS tools have been shown to approximate natural speech quality so closely that distinguishing synthetic from human speech is non-trivial~\cite{10356_181485, ijcai2024p0046}. To ensure that translation differences come from the MT systems rather than acoustic variations, we used consistent female voice presets across all synthesized audio. This approach reduces variability and is in line with previous works demonstrating that consistent speaker characteristics improve the reliability of MT system evaluation~\cite{10314895}.

% \begin{table*}[t]
% \centering
% \begin{tabular}{l|cc|cc}
% \hline
% \multirow{2}{*}{system} & \multicolumn{2}{c|}{English $\rightarrow$ German} & \multicolumn{2}{c}{English $\rightarrow$ Russian} \\
% & \textit{news} & idioms & \textit{news} & idioms \\
% \hline
% \multicolumn{5}{l}{\textit{Whisper Audio Encoder}} \\
% Whisper (Direct SLT) & \textit{0.8437} & 0.6402 & \textit{0.8318} & 0.6916 \\
% Whisper (ASR) $\rightarrow$ NLLB & \textit{0.8767} & 0.6774 & \textit{0.8523} & 0.7180 \\
% Whisper (ASR) $\rightarrow$ Seamless (MT) & \textit{0.8805} & 0.6703 & \textit{0.8603} & 0.7147 \\
% Whisper (ASR) $\rightarrow$ LLaMA & \textit{0.8685} & \textbf{0.6875} & \textit{0.8438} & \textbf{0.7339} \\
% \hline
% \multicolumn{5}{l}{\textit{Seamless M4T Audio Encoder}} \\
% Seamless (Direct SLT) & \textit{0.8697} & 0.6483 & \textit{0.8512} & 0.6941 \\
% Seamless (ASR) $\rightarrow$ NLLB & \textit{0.8672} & 0.6790 & \textit{0.8594} & 0.7025 \\
% Seamless (ASR) $\rightarrow$ Seamless (MT) & \textit{0.8729} & 0.6719 & \textit{0.8614} & 0.7185 \\
% Seamless (ASR) + LLaMA & \textit{0.8624} & \textbf{0.6871} & \textit{0.8454} & \textbf{0.7283} \\
% \hline
% \multicolumn{5}{l}{\textit{Text MT (upper bound performance)}} \\
% Seamless (Text MT) & \textit{0.8870} & 0.6784 & \textit{0.8694} & 0.7262 \\
% NLLB & \textit{0.8841} & 0.6749 & \textit{0.8664} & 0.7214 \\
% LLaMA & \textit{0.8724} & \textbf{0.6971} & \textit{0.8211} & \textbf{0.7354} \\
% \hline
% \end{tabular}
% \caption{Performance comparison of different translation systems}
% \label{tab:translation_performance}
% \end{table*}

\begin{table*}
\setlength{\tabcolsep}{0.5em}
\centering
\begin{tabular*}{\textwidth}{p{6.3cm}>{\columncolor{gray!20}}p{2cm} p{2cm}>{\columncolor{gray!20}}p{2cm} p{2cm}}
\toprule
\multirow{2}{*}{system} & \multicolumn{2}{c}{German $\rightarrow$ English} & \multicolumn{2}{c}{ Russian $\rightarrow$ English} \\
\cmidrule(lr){2-3} \cmidrule(lr){4-5}
& \textit{news} & idioms & \textit{news} & idioms \\
\midrule
\multicolumn{5}{l}{\textit{Whisper Audio Encoder}} \\
Whisper (Direct SLT) & \textit{0.8437} & 0.6402 & \textit{0.8318} & 0.6916 \\
Whisper (ASR) $\rightarrow$ NLLB & \textit{0.8767} & 0.6774 & \textit{0.8523} & 0.7180 \\
Whisper (ASR) $\rightarrow$ Seamless (MT) & \textit{0.8805} & 0.6703 & \textit{0.8603} & 0.7147 \\
Whisper (ASR) $\rightarrow$ LLaMA & \textit{0.8685} & 0.6875 & \textit{0.8438} & 0.7339 \\
Whisper (ASR) $\rightarrow$ DeepSeek & \textit{0.8887} & \textbf{0.7584} & \textit{0.8607} & \textbf{0.7873} \\
\midrule
\multicolumn{5}{l}{\textit{Seamless M4T Audio Encoder}} \\
Seamless (Direct SLT) & \textit{0.8697} & 0.6483 & \textit{0.8512} & 0.6941 \\
Seamless (ASR) $\rightarrow$ NLLB & \textit{0.8672} & 0.6790 & \textit{0.8594} & 0.7025 \\
Seamless (ASR) $\rightarrow$ Seamless (MT) & \textit{0.8729} & 0.6719 & \textit{0.8614} & 0.7185 \\
Seamless (ASR) $\rightarrow$ LLaMA & \textit{0.8624} & 0.6871 & \textit{0.8454} & 0.7283 \\
Seamless (ASR) $\rightarrow$ DeepSeek & \textit{0.8857} & \textbf{0.7635} & \textit{0.8667} & \textbf{0.7804} \\
\midrule
\multicolumn{5}{l}{\textit{Text MT (upper bound performance)}} \\
Seamless (Text MT and LLM) & \textit{0.8870} & 0.6784 & \textit{0.8694} & 0.7262 \\
NLLB & \textit{0.8841} & 0.6749 & \textit{0.8664} & 0.7214 \\
LLaMA & \textit{0.8724} & 0.6971 & \textit{0.8211} & 0.7354 \\
DeepSeek & \textit{0.8940} & \textbf{0.7675} & \textit{0.8741} & \textbf{0.7939} \\
\bottomrule
\end{tabular*}
\caption{Performance comparison of translation systems across modalities and approaches, showing COMET scores for both news and idiomatic content in German$\rightarrow$English and Russian$\rightarrow$English translation.}
\label{tab:translation_performance}
\end{table*}
\subsection{Evaluation Procedure}
To assess model performance, we employed both automatic and manual evaluation methods.
\subsubsection{Automatic Metrics}
For the automatic evaluation of translation quality, we utilize the COMET metric~\cite{rei-etal-2020-comet} of version \texttt{Unbabel/wmt22-comet-da}. COMET is a state-of-the-art framework that has shown a high correlation with human judgments. It assesses translations based on semantic equivalence and fluency.  This is particularly critical for idioms where literal translation fails to convey semantic equivalence, and contextual understanding is essential~\cite{song-xu-2024-benchmarking}.
By using COMET, we were able to ensure that both the intended meaning and the naturalness of idioms rather than form similarity are prioritized.

\subsubsection{Human Annotation of Translation Output}
To supplement COMET evaluations, two human annotators evaluated a random sample of 50 translations from each language-dataset-model combination using the annotation scheme in Table~\ref{tab:annotation_scheme}, where categories range from \emph{Correct} to \emph{Empty/Ellipsis}. For clear comparison, only encoder-decoder models were used for this evaluation. For idioms, annotators explicitly judged if figurative meaning was maintained by annotating correct translations as either \emph{Correct (Idiomatic)} or \emph{Correct (Paraphrase)}. The category \emph{Literal Translation} was also only used in idiom translation evaluation. The annotators resolved any disagreements through discussion to ensure consistent evaluation criteria.

\section{Results and Discussion}
\label{sec:results}
\subsection{Overall Performance}

Table~\ref{tab:translation_performance} presents COMET scores for German and Russian,  comparing model performance on news vs.\ idiom datasets. For each model, we further evaluated the differences in performance on two datasets using the Mann–Whitney U test. After applying Bonferroni correction for multiple comparisons, all models demonstrated statistically significant differences in performance on news vs.\ idioms with corrected 
\emph{p}-values below 0.001 for both language pairs. Additional statistical analyses, i.e. Kruskal-Wallis tests, standard deviation, and median performance comparisons, are provided in Appendix~\ref{sec:pairwise_comparison}. The word error rates of ASR transcription used in the cascaded systems are provided in Appendix~\ref{sec:wer}.

\textbf{MT and LLM vs.\ SLT:} The DeepSeek model largely outperforms all other models, especially on idiom translation. Other text-based systems (including NLLB, SeamlessM4T, and LLaMA variants) consistently outperform SLT systems (SeamlessM4T and Whisper) on idiom dataset regardless of language, and only in some cases on news dataset, such as NLLB and M4T with higher COMET scores for both German and Russian.

\textbf{Performance Drop on Idioms:} SLT systems' COMET scores decline sharply when moving from news to idioms (e.g., a 24.2\% drop from 0.844 to 0.640 in German$\rightarrow$English for Whisper).

\textbf{Cascaded Systems:} Although cascaded systems do not reach the end-to-end text-based systems' performance level, they mostly outperform end-to-end SLT systems. This seems to suggest that SLT systems errors are not solely due to ASR transcription but also reflect deeper challenges in the end-to-end systems. Such challenges may involve the integration of acoustic and semantic information, which is particularly important for semantically complex idiomatic language.

\subsection{Translation Error Distributions}
\label{sec:translation_categories}
Figure~\ref{fig:output_categories} displays the distribution of translation error categories (listed in Table~\ref{tab:annotation_scheme}) for each encoder-decoder model in the Russian$\rightarrow$English (top panel) and German$\rightarrow$English (bottom panel) translations. Two SLT systems (Whisper and SeamlessM4T) and two MT systems (NLLB and SeamlessM4T) were analyzed. As shown in Figure~\ref{fig:output_categories}, there is a clear difference in performance on news and idiom datasets. For news, both SLT and MT systems produce predominantly correct outputs. By contrast, idiomatic datasets see less correct and more divergent outputs. SLT and MT systems both produce a high proportion of the \textit{Literal Translation} category for idiom translation. This points to a shared challenge of recognizing idioms, although it is less pronounced in MT systems. Additionally, SLT systems are more likely to generate not only literal but also partially correct translations, while MT systems demonstrate a better, though far from perfect, handling of figurative language. These results emphasize the general shortfall of translation systems in capturing idiomatic meaning. These patterns emphasize the broader challenge that idiomatic expressions pose for current translation systems, revealing fundamental limitations in their ability to capture non-literal meaning.

\section{Layer-wise Analysis with DecoderLens}
To understand where translation systems capture or lose idiomatic meaning, we analyzed four encoder-decoder translation systems using DecoderLens \cite{langedijk2024decoderlenslayerwiseinterpretationencoderdecoder}: two SLT systems (Whisper and SeamlessM4T) and two MT systems (NLLB and SeamlessM4T).

DecoderLens enables analysis of intermediate representations by replacing the final encoder output with activations from each encoder layer, allowing the decoder to attend to these intermediate states. It reveals how semantic meaning evolves through the network by converting hidden representations into human-readable text.
For each model, we extracted outputs from all encoder layers and generated translations of 50 examples, which then were annotated by two human annotators using the scheme in Table~\ref{tab:annotation_scheme}. The results highlight key differences between SLT and MT systems in processing figurative language.

\subsection{Results of Layer-wise Analysis with DecoderLens}
\label{sec:decoderlens_analysis}

Table~\ref{tab:decoderlens_condensed} presents an example of  layer-by-layer English translation outputs from Whisper SLT system via DecoderLens for a Russian idiomatic item. As shown in the example, Layers~0--20 consistently produce empty or punctuation-only strings, indicating that the model has yet to form a meaningful textual output. Starting from Layer~21, the system attempts to generate text but mostly produces \emph{Hallucinations}. Only in the last few layers does the system start to align with the original text (reflected by the \emph{Incorrect but Relevant} category), and eventually produce a \emph{Partially Correct} output at Layer~31. However, Layer~32 only manages to output a \emph{Literal Translation}, further showing that the model fails to preserve the figurative sense of the idiom 'still waters.'

\vspace{2mm}
\begin{table*}[t!]
\centering
\small
\renewcommand{\arraystretch}{1.1}
\rowcolors{2}{gray!10}{white}
\setlength{\tabcolsep}{6pt}

\begin{tabular}{l p{9.1cm} p{3.7cm}}
\toprule
\rowcolor{gray!30}
\textbf{Layer(s)} & \textbf{Output (DecoderLens)} & \textbf{Annotation}\\ 
\midrule

\textbf{0--1} & \emph{(empty)} & Empty/Ellipsis \\
\textbf{2--10} & \emph{(empty)} & Empty/Ellipsis \\
\textbf{11}  & \texttt{,,,,,,,,,,,,,,,,,,,,,,,,,,,,,,,,,,,,,,,,,,,,,,,,,,,,,,,,,,,,,,,,,,,,,,} & Empty/Ellipsis \\
\textbf{12--20} & \emph{(empty)} & Empty/Ellipsis \\

\textbf{21} & The world is full of people. & Incorrect (Hallucination) \\
\textbf{22} & The first part of the video is about the first part of the video. & Incorrect (Hallucination) \\
\textbf{23} & The next day. & Incorrect (Hallucination) \\
\textbf{24} & The story of the story of the story of the story of the story of \dots (truncated) & Incorrect (Hallucination) \\
\textbf{25} & The story of the man who was killed by a woman. & Incorrect (Hallucination) \\
\textbf{26} & The T-shirt of the Russian national team. & Incorrect (Hallucination) \\
\textbf{27--28} & The End & Incorrect (Hallucination) \\
\textbf{29} & You know, I'm a man of my word. & Incorrect (Relevant) \\
\textbf{30} & In the quiet you know who is the best. In the quiet you know who is the best. & Incorrect (Relevant) \\
\textbf{31} & In the silent moment you know who is in charge. The appearance and manner of a person can be deceiving. & Partially Correct \\
\textbf{32} & In a quiet room you know who is leading. The appearance and manner of human behavior can be deceiving. & Literal Translation \\

\bottomrule
\end{tabular}
\caption{
\textbf{DecoderLens Layer-by-layer Outputs: Whisper for a Russian Idiomatic Sentence.}\\
\textit{Original (Russian):} \selectlanguage{russian}{``В тихом омуте сам знаешь кто водится… Внешность и манера поведения человека бывают обманчивы.''}\\\selectlanguage{english}
\textit{Gold translation (English):} ``You know what they say about still waters. A person's appearance and behavior can be deceptive.''
}
\label{tab:decoderlens_condensed}
\end{table*}

Figure~\ref{fig:decoderlens_categories} shows a layer-by-layer breakdown of translation outputs for 50 examples of each evaluated system using the DecoderLens method, based on the categories introduced in Table~\ref{tab:annotation_scheme}. Each subplot corresponds to a particular system and domain (news vs.\ idioms) for Russian and German. %I have problems when add a comment, so I put a comment here: you sometimes use "system" and sometimes use "model" referring to the, you know, the models, is there a clear distinction? If so, I suggest to make it clearer, and if not, choose one and consistently use the chosen one.

For all the available data, direct SLT systems start to produce meaningful translations only in higher encoder layers. From there, they gradually improve from producing \emph{Partially Correct} outputs to \emph{Paraphrased}, \emph{Literal}, and \emph{Correct} translations in the final layers. For relatively straightforward news text, the model progressively refines its representations towards correct translations.  
By contrast, when translating idioms, SLT systems are more prone to literal translations, with only minor improvements in higher layers. 

MT systems also have difficulties moving away from literal translations for idiomatic inputs. In general, however, their transitions across layers are smoother, which indicates a different internal strategy for capturing semantics.

\subsection{Cross-Language Differences}
Although both German and Russian see drops in idiomatic performance, German has a larger gap (0.198 on average) between news and idioms, while Russian’s gap is around 0.143. The ranking of systems, however, is mostly consistent across the two languages.

\section{Conclusion}
\label{sec:conclusion}
In this work, we provide a systematic comparison of speech-to-text (SLT), text-to-text machine translation (MT), and Large Language Models (LLMs) when challenged with idiomatic datasets in German$\rightarrow$English and Russian$\rightarrow$English. Our findings reveal the following: % I suggest not using bullets/numerates for conclusion.

\begin{enumerate}
    \item \textbf{SLT underperforms for idioms.} 
Both SLT and MT systems struggle with idiomatic translation, as reflected by performance drops of COMET scores on idiom dataset compared to conventional news. Notably, the performance gap between news and idiom datasets is more pronounced for SLT systems, while MT and LLMs are better at translating idioms.

    \item \textbf{Layer-wise analysis highlights structural differences between speech-to-text and text-to-text systems.} 
    Using DecoderLens, we observe that SLT systems only start to 'refine' their translation output in the higher layers, and revert to literal translation more frequently even in higher encoder layers. MT systems, on the other hand, show a gradual improvement in capturing the intended sense when moving from intermediate to higher layers.
\end{enumerate}

Overall, our study shows that translating idioms remains a bigger challenge for SLT systems compared to MT, LLMs, and cascaded systems. Although SeamlessM4T and Whisper perform competitively on conventional news, cascaded approaches combining strong ASR and text-based components provide better handling of figurative language, likely due to text-based systems' stronger semantic processing. These findings highlight the need for idiom-specific strategies and improved representations of idioms in SLT systems. For practical applications, we recommend using cascaded systems when translating speech likely to contain idiomatic expressions. We hope this study will inspire further research on figurative language in speech translation.

\section*{Ethical statement}
The annotators involved in this study were compensated for their work on hourly basis.

\section*{Limitations}
Annotating translation output is inherently subjective. Morever, our approach focuses only on translating German and Russian to English, while idiom usage varies widely across languages. The use of synthetic speech may differ from real-world spontaneous speech though prior work suggests minimal impact on core translation errors. Finally, DecoderLens analysis is limited to encoder-decoder architectures and may not capture idiom handling in purely decoder-based systems like LLaMA.

%\section{Preamble}

\section*{Acknowledgments}
We thank Mutahar Aamir for his help with the data annotation, as well as the anonymous reviewers for their constructive feedback. This research is funded by the Deutsche Forschungsgemeinschaft (DFG, German Research Foundation), Project-ID 232722074 – SFB 1102.
\bibliography{custom}

\appendix

\section{Appendix A}
\label{sec:appendix_A}
\subsection{Prompt for LLaMA~3 fine-tuned for Russian} 
\begin{tcolorbox}[colframe=black!75, colback=white!5, sharp corners, boxrule=0.5mm]
You are a professional translator who translates from Russian to English. \\
Only generate the target sentence, and nothing else. Follow the example below: \\

Input sentence: \selectlanguage{russian}{У меня нет воды.} \\
Translation: I don't have water. \\

Translate this sentence:
\end{tcolorbox}

\subsection{Prompt for LLaMA~3 fine-tuned for German}
\begin{tcolorbox}[colframe=black!75, colback=white!5, sharp corners, boxrule=0.5mm]
You are a professional translator who translates from German to English. \\
Only generate the target sentence, and nothing else. Follow the example below: \\

Input sentence: Ich habe kein Wasser. \\
Translation: I don't have water. \\

Translate this sentence:
\end{tcolorbox}
\hfill \break

\section{Appendix B}
\label{sec:pairwise_comparison}

\begin{table*}[htbp]
\centering
\caption{Performance analysis of translation models using COMET scores for German$\rightarrow$English data}
\label{tab:detailed-model-comparison}
\small
\begin{subtable}{0.45\textwidth}
\centering
\caption{\textbf{German News}: COMET score analysis for German$\rightarrow$English translation on news data}
\label{tab:detailed-model-comparison}
\small
\begin{tabular}{lrrr}
\toprule
\textbf{Model} & \textbf{Mean} & \textbf{Median} & \textbf{Std} \\
\midrule
DeepSeek & \textbf{0.894} & 0.901 & 0.054 \\
Whisper + DeepSeek & 0.889 & 0.896 & 0.055 \\
M4T Text & 0.887 & 0.894 & 0.059 \\
M4T ASR + DeepSeek & 0.886 & 0.892 & 0.055 \\
NLLB & 0.884 & 0.898 & 0.078 \\
Whisper + M4T & 0.880 & 0.889 & 0.062 \\
Whisper + NLLB & 0.877 & 0.894 & 0.083 \\
M4T ASR + MT & 0.873 & 0.883 & 0.066 \\
LLaMA & 0.872 & 0.885 & 0.062 \\
M4T Audio & 0.870 & 0.879 & 0.065 \\
Whisper + LLaMA & 0.869 & 0.882 & 0.067 \\
M4T ASR + NLLB & 0.867 & 0.884 & 0.084 \\
M4T ASR + LLaMA & 0.862 & 0.873 & 0.066 \\
Whisper & 0.844 & 0.854 & 0.074 \\
\midrule
\multicolumn{4}{l}{\textit{Statistical Analysis:}} \\
\multicolumn{4}{l}{Kruskal-Wallis H = 179.09} \\
\multicolumn{4}{l}{$p$-value < 2.60 $\times$ 10\textsuperscript{-31}} \\
\bottomrule
\end{tabular}
\end{subtable}
\hfill
\begin{subtable}{0.45\textwidth}
\centering
\caption{\textbf{German Idioms}: COMET score analysis for German$\rightarrow$English translation on idiomatic data}
\label{tab:detailed-model-comparison}
\small
\begin{tabular}{lrrr}
\toprule
\textbf{Model} & \textbf{Mean} & \textbf{Median} & \textbf{Std} \\
\midrule
DeepSeek & \textbf{0.767} & 0.779 & 0.128 \\
M4T ASR + DeepSeek & 0.764 & 0.759 & 0.131 \\
Whisper + DeepSeek & 0.758 & 0.758 & 0.133 \\
LLaMA & 0.697 & 0.698 & 0.136 \\
Whisper + LLaMA & 0.687 & 0.690 & 0.134 \\
M4T ASR + LLaMA & 0.687 & 0.692 & 0.138 \\
M4T ASR + NLLB & 0.679 & 0.682 & 0.132 \\
M4T Text & 0.678 & 0.684 & 0.131 \\
Whisper + NLLB & 0.677 & 0.684 & 0.130 \\
NLLB & 0.675 & 0.665 & 0.130 \\
M4T ASR + MT & 0.672 & 0.670 & 0.133 \\
Whisper + M4T & 0.670 & 0.676 & 0.132 \\
M4T Audio & 0.648 & 0.644 & 0.125 \\
Whisper & 0.640 & 0.639 & 0.124 \\
\midrule
\multicolumn{4}{l}{\textit{Statistical Analysis:}} \\
\multicolumn{4}{l}{Kruskal-Wallis H = 275.74} \\
\multicolumn{4}{l}{$p$-value < 2.82 $\times$ 10\textsuperscript{-51}} \\
\bottomrule
\end{tabular}
\end{subtable}
 \textit{Note:} Models are sorted by mean COMET score. The Kruskal-Wallis test indicates statistically significant differences between model performances. The best-performing models (DeepSeek) is shown in bold.
\end{table*}

\begin{table*}[htbp]
\centering
\caption{Performance analysis of translation models using COMET scores for Russian$\rightarrow$English data}
\label{tab:detailed-model-comparison}
\small

\begin{subtable}{0.45\textwidth}
\centering
\caption{\textbf{Russian News}: COMET score analysis for Russian$\rightarrow$English translation on news data}
\label{tab:detailed-model-comparison}
\small
\begin{tabular}{lrrr}
\toprule
\textbf{Model} & \textbf{Mean} & \textbf{Median} & \textbf{Std} \\
\midrule
DeepSeek & \textbf{0.874} & 0.878 & 0.051 \\
M4T Text & 0.869 & 0.874 & 0.054 \\
M4T ASR + DeepSeek & 0.867 & 0.871 & 0.054 \\
NLLB & 0.866 & 0.873 & 0.056 \\
M4T ASR + MT & 0.861 & 0.866 & 0.059 \\
Whisper + DeepSeek & 0.861 & 0.872 & 0.078 \\
Whisper + M4T & 0.860 & 0.868 & 0.063 \\
Whisper + NLLB & 0.859 & 0.868 & 0.064 \\
M4T ASR + NLLB & 0.852 & 0.864 & 0.068 \\
M4T Audio & 0.851 & 0.858 & 0.060 \\
M4T ASR + LLaMA & 0.845 & 0.851 & 0.061 \\
Whisper + LLaMA & 0.844 & 0.851 & 0.067 \\
Whisper & 0.832 & 0.836 & 0.070 \\
LLaMA & 0.821 & 0.858 & 0.122 \\
\midrule
\multicolumn{4}{l}{\textit{Statistical Analysis:}} \\
\multicolumn{4}{l}{Kruskal-Wallis H = 127.89} \\
\multicolumn{4}{l}{$p$-value < 5.49 $\times$ 10\textsuperscript{-21}} \\
\bottomrule
\end{tabular}
\end{subtable}
\hfill
\begin{subtable}{0.45\textwidth}
\centering
\caption{\textbf{Russian Idioms}: COMET score analysis for Russian$\rightarrow$English translation on idiomatic data}
\label{tab:detailed-model-comparison}
\small
\begin{tabular}{lrrr}
\toprule
\textbf{Model} & \textbf{Mean} & \textbf{Median} & \textbf{Std} \\
\midrule
DeepSeek & \textbf{0.794} & 0.801 & 0.084 \\
Whisper + DeepSeek & 0.787 & 0.794 & 0.090 \\
M4T ASR + DeepSeek & 0.780 & 0.791 & 0.093 \\
LLaMA & 0.735 & 0.741 & 0.105 \\
Whisper + LLaMA & 0.734 & 0.737 & 0.103 \\
M4T ASR + LLaMA & 0.728 & 0.734 & 0.108 \\
M4T Text & 0.726 & 0.734 & 0.108 \\
NLLB & 0.721 & 0.736 & 0.117 \\
M4T ASR + MT & 0.718 & 0.726 & 0.111 \\
Whisper + NLLB & 0.718 & 0.735 & 0.115 \\
Whisper + M4T & 0.715 & 0.719 & 0.110 \\
M4T ASR + NLLB & 0.703 & 0.710 & 0.118 \\
M4T Audio & 0.694 & 0.699 & 0.116 \\
Whisper & 0.692 & 0.690 & L0.106 \\
\midrule
\multicolumn{4}{l}{\textit{Statistical Analysis:}} \\
\multicolumn{4}{l}{Kruskal-Wallis H = 276.88} \\
\multicolumn{4}{l}{$p$-value < 1.62 $\times$ 10\textsuperscript{-51}} \\
\bottomrule
\end{tabular}
\end{subtable}
 \textit{Note:} Models are sorted by mean COMET score. The Kruskal-Wallis test indicates statistically significant differences between model performances. The best-performing model (DeepSeek) is shown in bold.
\end{table*}
\break
\section{Appendix C}
\label{sec:wer}
\begin{table}[ht]
\centering
\begin{tabular}{lcccc}
\hline
\textbf{System} & \multicolumn{2}{c}{\textbf{German}} & \multicolumn{2}{c}{\textbf{Russian}} \\
                & News & Idioms & News & Idioms \\
\hline
Whisper         & 0.040 & 0.024 & 0.081 & 0.084 \\
Seamless     & 0.085 & 0.037 & 0.124 & 0.145 \\
\hline
\end{tabular}
\caption{Average Word Error Rate of automatic speech recognition used in cascaded translation systems. Lower is better.}
\label{tab:wer_summary}
\end{table}

\end{document}